\documentclass{edm_article}
\usepackage{balance}
\usepackage{enumitem}
\usepackage{booktabs}
\usepackage{multirow}
\usepackage{graphicx}
\usepackage{balance}
\usepackage[caption=false]{subfig}
\usepackage{xcolor}
\usepackage{tablefootnote}

\begin{document}

\title{Evaluating the Explainers: Black-Box Explainable Machine Learning for Student Success Prediction in MOOCs}

\numberofauthors{5}
\author{
% 1st. author
\alignauthor
Vinitra Swamy\\
        \affaddr{EPFL}\\
        \email{\normalsize vinitra.swamy@epfl.ch}
% 2nd. author
\alignauthor
Bahar Radmehr\\
        \affaddr{Sharif University}\\
        \email{\normalsize radmehr.bahar@ee.sharif.edu}
% 3rd. author
\alignauthor
Natasa Krco\\
        \affaddr{EPFL}\\
        \email{\normalsize natasa.krco@epfl.ch}
\and  % use '\and' if you need 'another row' of author names
% 4th. author
\alignauthor Mirko Marras\\
        \affaddr{University of Cagliari}\\
        \email{\normalsize mirko.marras@acm.org}
% 5th. author
\alignauthor Tanja Käser\\
        \affaddr{EPFL}\\
        \email{\normalsize tanja.kaeser@epfl.ch}
}

\maketitle

\begin{abstract}
Neural networks are ubiquitous in applied machine learning for education. Their pervasive success in predictive performance comes alongside a severe weakness, the lack of explainability of their decisions, especially relevant in human-centric fields. We implement five state-of-the-art methodologies for explaining black-box machine learning models (LIME, PermutationSHAP, KernelSHAP, DiCE, CEM) and examine the strengths of each approach on the downstream task of student performance prediction for five massive open online courses. Our experiments demonstrate that the families of explainers \textbf{do not agree} with each other on feature importance for the same Bidirectional LSTM models with the same representative set of students. We use Principal Component Analysis, Jensen-Shannon distance, and Spearman's rank-order correlation to quantitatively cross-examine explanations across methods and courses. Furthermore, we validate explainer performance across curriculum-based prerequisite relationships. Our results come to the concerning conclusion that the choice of explainer is an important decision and is in fact paramount to the interpretation of the predictive results, even more so than the course the model is trained on. Source code and models are released at \texttt{\small http://github.com/epfl-ml4ed/evaluating-explainers}.

\end{abstract}


\section{Introduction}

The steep rise in popularity of neural networks has been closely mirrored by the adoption of deep learning for education. For the majority of educational data modeling tasks such as student success prediction (e.g., \cite{imran2019predicting}), estimating early dropout (e.g., \cite{xing2019dropout}), and knowledge tracing (e.g., \cite{piech2015deep, abdelrahman2019knowledge}), the recent literature relies on neural networks to reduce human involvement in the pipeline and boost overall prediction accuracy. Unfortunately, these advances come at a significant cost: traditional machine learning techniques (e.g., linear regression, SVMs, decision trees) are simple, but interpretable, where deep learning techniques trade transparency for the ability to capture complex data representations \cite{molnar2022}.

There is a compelling need for interpretability in models dealing with human data, especially in education. \cite{webb2021machine} emphasizes that explainability and accountability should be incorporated in machine learning system design to meet social, ethical and legislative requirements. Other work \cite{conati2018ai} strongly argues for the necessity of interpretable models in education, specifically in settings where students can see the effect of a decision but not the reasoning behind it (e.g., Open Learner Models). Predictions of student performance are often used to determine underachieving students for targeted downstream interventions. Identifying important features motivating failure or dropout predictions is crucial in designing effective, personalized interventions.

However, there exists only a handful of papers focusing on explainability in the field of machine learning for education. For example, \cite{lu2020towards} examined the inner workings of deep learning models for knowledge tracing through layer-relevance propagation. Other researchers \cite{hasib2022lime} experimented with traditional machine learning models for student success prediction and implemented local explanations with LIME for transparency in the best performing model. Additionally, \cite{baranyi2020interpretable} used SHAP feature importances to interpret student dropout prediction models. \cite{mu2020towards} suggested interventions for wheel-spinning students based on Shapley values. Finally, \cite{vultureanu2021improving} explored LIME on ensemble machine learning methods for student performance prediction, \cite{scheers2021interactive} integrated LIME explanations in student advising dashboards, and \cite{pei2021} used LIME for interpreting models identifying at-risk students.

While field of neural network explainability is also nascent in the broader machine learning community, the last five years have shown a sharp increase in research and industry interest in this topic. Local, instance-based explainability methods like LIME \cite{lime} and SHAP \cite{shap} have become immensely popular. These methods have been successfully applied on models predicting ICU mortality \cite{katuwal2016machine}, non-invasive ventilation for ALS patients \cite{ferreira2021predictive}, and credit risk \cite{creditrisk}. Recent work in counterfactual explanations \cite{dice, alibi, cem} searches for a minimal subset of features that leads to the prediction alongside a minimal feature subset that needs to be changed for the prediction to change. Counterfactuals have been used in tasks like image classification \cite{goyal2019counterfactual}, loan repayment \cite{pawelczyk2020learning}, and grouping websites into topics for safe-advertising \cite{martens2014explaining}.

Although the explainability corpora is growing, there is a clear gap in explainability literature for education, with an even more pressing need for work (quantitatively) comparing different explainability methods. To the best of our knowledge, current research on explainability in education is exclusively applied: the majority of previous research implements only one specific explainability method to interpret the predictions of their proposed approach.

% 3. What is the gap?
% - Not really directly needed here - it is kind of clear from 2

% 4. What have we done:
% - In this paper, we aim to...
% - We suggest this  method
% - We evaluate 5 methods on 5 courses to answer the following 3 research questions

% 5. Our results demonstrate...

To address this research gap, we examine and compare five popular instance-based explainability methods on student success prediction models for five different massive open online courses (MOOCs). We formulate comparable feature importance scores for each explainer, scaled between $[0,1]$ on a uniformly sampled, stratified representative set of students. To quantitatively compare the feature importance distributions, we propose the use of different measures: rank-based metrics (Spearman's rank-order correlation), distance metrics (Jensen Shannon Distance), and dimensionality analysis (Principal Component Analysis). We validate the explanations through an analysis of feature importance on a MOOC with known prerequisite relationships in the underlying curriculum. With our experiments, we address three research questions: 1) How similar are the explanations of different explainability methods for a specific course (\textbf{RQ1})? 2) How do explanations (quantitatively) compare across courses (\textbf{RQ2})? 3) Do explanations align with prerequisite relations in a course curriculum (\textbf{RQ3})?

Our results demonstrate that the feature importance distributions extracted by different explainability methods for the same model and course differ significantly from each other. When comparing the feature importances across courses, we see that LIME is far apart from all other methods due to selecting a sparse feature set. Furthermore, our findings show that the choice of explainability method influences the feature importance distribution much more than the course the model is predicting on. Our examination on prerequisite relationships between features further indicates that the three families of methods are only partially able to uncover prerequisite dependencies between course weeks. Source code and models are released on Github\footnote{ \texttt{\small http://github.com/epfl-ml4ed/evaluating-explainers}}.

\section{Methodology}

\begin{figure*}[h!]
\centering
\includegraphics[width=\linewidth, trim=0 4 0 4,clip]{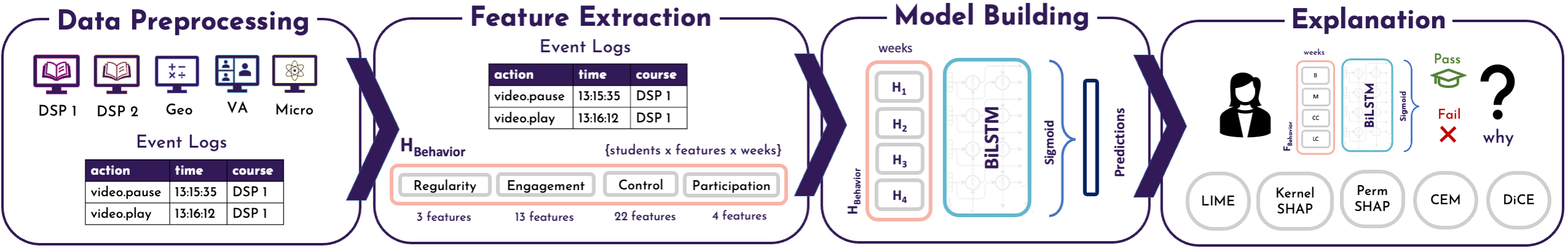}
\label{fig:pipeline}
\caption{Our experimental pipeline, from data processing to post-hoc explainability methods.}
\end{figure*}

The goal of this paper is to compare explanations from deep learning models tasked with identifying student success prediction in MOOCs. In this section, we formalize the student success prediction task addressed in this paper including the data collection and preprocessing, feature extraction, and model preparation. We then introduce the considered explainability methods and describe the process to extract explanations for student success predictions from a trained model, showcased as feature importance weights.

\subsection{Formal Preliminaries}
We consider a set of students $S$ enrolled in a course $c$ part of an online educational offering $\mathcal C $. Course $c$ has a predefined weekly schedule consisting of $N = |\mathbb{O}|$ learning objects from a catalog $\mathbb{O}$. Students enrolled in a course interact with the learning objects included in the course schedule, generating a time-wise clickstream (e.g., a sequence of video plays and pauses, quiz submissions). We denote a clickstream in a course $c$ for a student $s \in S$ as a time series $I_s = \{i_1, \ldots, i_{K_s}\}$ with $K_s$ being the total number of interactions of student $s$ in course $c$. Each interaction $i \in I_s$ is represented by a tuple $(t, a, o)$, including a timestamp $t$, an action $a$ (videos: load, play, pause, stop, seek, speed; quiz: submit), and a learning object $o \in \mathbb{O}$ (video, quiz). Given the weekly course schedule, we assume that $t_w$ identifies the time $t$ where the course week $w \in \{0, \dots, W\}$ ends, and that the clickstream of student $s$ generated until the end of the $w$ week can be denoted as $I^{t_w}_s$. We also assume that the course schedule includes one or more assignments per week and that the grade record of student $s$ across course assignments is denoted as $G_s = [g_1, \dots, g_W]$, where $g_w \in G_s$ is the grade student $s$ received on the assignment in week $w$. In the case of multiple graded assignments for a certain week, we considered the average score of graded assignments for that week and scored non-attempted assignments with $0$. We denote as $y_s \in \{0, 1\}$ the success label for student $s$.

\subsection{Data Preprocessing}
\label{sec:log-preproc}

A significant portion of MOOC students enroll just to watch a few videos or find that the curriculum material is not what they expected and drop out of the course in the first weeks \cite{onah2014dropout, goopio2021mooc}. It follows that it is easy to predict the success labels $y_s$ for this selection of students by simply looking at their initial few weekly assignment grades in $G_s$. Therefore, optimizing complex deep learning models for predicting student success on early-dropout students is inefficient. Using these complex deep models also leads to less interpretable predictions in comparison with traditional models. For this subset of early-dropout students, traditional models can both achieve a comparable accuracy and still remain interpretable. To identify early-dropout students, we fit a \textit{Logistic Regression} model on the assignment grades of the first two course weeks. The input data is the vector $G_s^W$, where $W$ is the number of course weeks ($W=2$ in our experiments) whereas the ground truth is the student success label $y_s$. Once the model is fitted, we filter out the students that had a predicted probability of course failure $\hat{p}_{s}>0.99$. We determine the optimal threshold via a grid search over $\{0.96, 0.97, \ldots, 0.999\}$, maximizing the model balanced accuracy. Henceforth, we consider $S$ to be the student population obtained after the early-dropout student filtering.

\subsection{Feature Extraction}
As an input for our student success prediction models, we consider a set of behavioral features extracted for each student $s \in S$ based on their interactions $I_s$. We include four feature sets proved to have high predictive power for success prediction in MOOCs \cite{marras2021can}. Given the size and variety of the course data considered in our study, we included all features of the four features sets, instead of considering only the specific features identified as important by at least one course \cite{marras2021can}. Formally, given interactions $I_s$ generated by students $S$ until a course week $w$, we create a matrix $H \subset \mathbb{R}^{|S| \times w \times f}$ (i.e, each feature in the feature set is computed per student per week), where $f \in \mathbb{N}$ is the dimensionality of the feature set. We focus on the following behavioral aspects:

\begin{itemize}[leftmargin=*,nolistsep]
    \item \textbf{Regularity} features ($H_1$, shape: $|S| \times w \times 3$) monitor the extent to which a student follows regular study habits \cite{boroujeni2016quantify}.
    \item \textbf{Engagement} features ($H_2$, shape: $|S| \times w \times 13$) monitor the extent to which a student is engaged in the course \cite{chen2020utilizing}.
    \item \textbf{Control} features ($H_3$, shape: $|S| \times w \times 22$) measure the fine-grained video consumption per student \cite{lalle2020data}.
    \item \textbf{Participation} features ($H_4$, shape: $|S| \times w \times 4$)  monitor attendance on videos/quizzes based on the schedule \cite{marras2021can}.
\end{itemize}

\vspace{2mm} We extract the above features for each student $s$ and concatenate features across sets to obtain the final combined behavioral features $h_s$ per student. The overall matrix of features is defined as $H \in \mathbb{R}^{|S| \times w \times 42}$, with $H = [H_1 \cdot H_2 \cdot H_3 \cdot H_4]$ ($\cdot$ denotes a concatenation). Due to the different scales, we perform a min-max normalization per feature in $H$ (i.e., we scale the feature between 0 and 1 considering all students and weeks for that feature). We elaborate on the most important features later on in the paper as highlighted by the analyses in subsequent experiments (e.g., Table \ref{tab:features}).

\renewcommand{\arraystretch}{1.2}
\begin{table*}[]
\centering
\small
\begin{tabular}{@{}lll@{}}
\toprule
\textbf{Set} & \textbf{Feature} & \textbf{Description} \\ \midrule
\multirow{3}{*}{\textit{Regularity}} & DelayLecture & The average delay in viewing video lectures after they are released to students. \\
 & RegPeakTimeDayHour & Regularity peak based on entropy of the histogram of user's activity over time. \\
 & RegPeriodicityM1 & The extent to which the hourly pattern of user’s activities repeats over days. \\
 \midrule
\multirow{9}{*}{\textit{Engagement}} & AvgTimeSessions & The average of users' time between subsequent sessions. \\
 & NumberOfSessions & The number of unique online sessions the student has participated in. \\
 & RatioClicksWeekendDay & The ratio between the number of clicks in the weekend and the weekdays \\
 & StdTimeSessions & The standard deviation of users' time between subsequent sessions. \\
 & TotalClicksProblem & The number of clicks that a student has made on problems this week. \\
 & TotalClicksWeekend & The number of clicks that a student has made on the weekends. \\
 & TotalTimeProblem & The total (cumulative) time that a student has spent on problem events. \\
 & TotalTimeVideo & The total (cumulative) time that a student has spent on video events. \\
 & StdTimeBetweenSessions & The standard deviation of the time between sessions of each user. \\
 \midrule
\multirow{3}{*}{\textit{Control}} & AvgReplayedWeeklyProp & The ratio of videos replayed over the number of videos available. \\
 & AvgWatchedWeeklyProp & The ratio of videos watched over the number of videos available. \\
 & FrequencyEventLoad & The frequency between every Video.Load action and the following action. \\
 \midrule
\multirow{4}{*}{\textit{Participation}} & CompetencyAnticipation & The extent to which the student approaches a quiz provided in subsequent weeks. \\
 & ContentAlignment & The number of videos for that week that have been watched by the student. \\
 & ContentAnticipation & The number of videos covered by the student from those that are in subsequent weeks. \\
 & StudentSpeed & The average time passed between two consecutive attempts for the same quiz. \\
 \bottomrule
\end{tabular}
\caption{Features used in model explainability analysis. For brevity, we only list the $19$ features that have been identified as important by at least one explainability method in our analysis in Section \ref{sec:results}.}
\label{tab:features}
\end{table*}
\renewcommand{\arraystretch}{1.0}

\subsection{Model Building}
Given a course $c$, we are interested in creating a success prediction model that can accurately predict the success label $y_s$ for student $s$, given the extracted behavioral features $h_s$. To this end, we rely on a neural architecture based on \emph{Bidirectional LSTMs}, which can provide a good trade-off between effectiveness and efficiency\footnote{Experimental details can be found in Appendices A and B.}. The model input is represented by $H$, i.e., the extracted behavior features, having a shape of $|S| \times W \times 42$. NaN values were replaced with the minimum score the student can receive for each respective feature. These features are then fed into a neural architecture composed by two simple yet effective \emph{BiLSTM} layers of size 32 and 64  (loopback of $3$) and a \emph{Dense} layer (with Sigmoid activation) having a hidden size of $1$. The model outputs the probability the student will pass the course.

\subsection{Explanation}
\label{sec:explain}

Input behavioral features contribute with varying levels of importance to the prediction provided by a success prediction model. We unfortunately cannot examine the importance of these features directly, since deep neural networks act as \textbf{black boxes}. Explainability methods can therefore be adopted to approximate the contributions of each feature in $H$ towards the prediction associated with a specific student $s$. To explore this aspect, we consider five instance-based explainability methods that are popular in the literature and cover different method families \cite{linardatos2020explainable, molnar2022}. We then compute the feature importance vector for each student $s$, based on each explainability method. Formally, given an explainability method, we denote $e_s \in \mathbb R^{w*42}$ as the feature importance weights returned by the explainability method for student $s$. The feature importance weight $e_s[i]$ is a score, comparable across explainability methods, that represents the importance of feature $h_s[i]$ to the model's individual prediction for student $s$. The considered explainability methods are described below.

\textbf{LIME} \cite{lime} trains a local linear model to explain each individual student instance $h_s$. To this end, it first generates perturbed instances $h_s^{1}, h_s^{2} \ldots h_s^{n}$ by shifting the feature values of $h_s$ a small amount. These new instances are then passed to the original model to get their associated predictions. Finally, a local interpretable model (e.g., a Support Vector Machine) is trained on the perturbed instances (input) and the corresponding predictions obtained from the original model (labels), weighting perturbed instances by proximity to the original instance. Mathematically, the local model can be expressed with the following equation:
\begin{equation}
\text{LIME}(h_s) = \text{argmin}_{g'\in G'}L(g, g', \pi_{h_s})+\Omega(g')
\end{equation}
where $h_s$ is the instance being explained, $G'$ is the family of all possible explanations, $L$ the loss that measures how close the predictions of the explainer $g'$ are to the predictions of the original model $g$, $\pi_{h_s}$ is the feature proximity measure, and $\Omega(g')$ represents the complexity of the local model. As LIME returns feature weights $\pi_1 \ldots \pi_{|h_s|}$ representing the feature influence on the final decision, we consider these absolute values to be the importance scores $e_s$, and scale them to the interval $[0,1]$, where $1$ indicates high importance.

\textbf{KernelSHAP} \cite{shap} draws inspiration from game-theory based Shapley values (computing feature contributions to the resulting prediction) and LIME (creating locally interpretable models). This SHAP variant uses a specially-weighted local linear regression to estimate SHAP values for any model. Let $x = h_s$ be the student instance being explained. A point $x'$ in the neighborhood of $x$ is generated by first sampling a coalition vector $z \in \mathbb R^|h_s|$. The coalition vector uses a binary mask to determine which features from $x$ will be kept the same in the new instance $x'$, and which will be replaced by a random value from the data distribution of that feature in $H$. Feature importance weights for each new instance $x'$ are calculated using a predefined kernel, after which the local model can be trained. A SHAP explanation is mathematically defined as:
\begin{equation}
g'(z') = \pi_0 + \sum_{h_{s=1}}^{|h_s|} \pi_{h_s}z'_{h_s}
\end{equation}
where $g'$ is the local explainer, $\pi_{h_s}\in \mathbb{R}$ is the SHAP value (feature attribution) of feature $h_s$, and $z'\in\{0,1\}^{|h_s|}$ is the coalition binary value. To achieve Shapley compliant weighting, Lundberg et al. \cite{shap} propose the SHAP kernel:
\begin{equation}
\pi_{h_s}(z')=\frac{({|h_s|}-1)}{\left(\begin{array}{c}
{|h_s|} \\
|z^{\prime}|
\end{array}\right)\left|z^{\prime}\right|\left({|h_s|}-\left|z^{\prime}\right|\right)}
\end{equation}
where $|h_s|$ is the maximum coalition size and $|z'|$ is the number of features present in coalition instance $z'$ \cite{molnar2022}.

SHAP methods directly provide values $\pi_{h_s}$ representing the feature contribution to the prediction $y_s$ of instance $s$. To obtain the importance scores $e_s$, we apply the same transformation as LIME, by taking the absolute values of the SHAP feature attributions and scaling them to the interval $[0,1]$.

\textbf{PermutationSHAP (PermSHAP)} \cite{shap} is very similar to the KernelSHAP formulation, but does not require the tuning of a regularization parameter or a kernel function.  We made the decision to include both KernelSHAP and PermSHAP as a form of validation of our comparative evaluation analysis; the distance between two very similar SHAP methods is expected to be smaller than the distance between these SHAP methods and other families of explainability methods. PermSHAP approximates the Shapley values of features by iterating completely through an entire permutation of the features in both forward and reverse directions (antithetic sampling). To extract the feature importance vector $e_s$, we again consider the absolute values of the SHAP feature attributions and scale to the interval $[0,1]$.

\textbf{Contrastive Explanation Method (CEM)} \cite{cem} identifies which features need to be present (pertinent positives) or which features must be absent (pertinent negatives) in order to maintain the model prediction $y_s$ for a student $s$ with behavioral features $h_s$ \cite{cem}. For our setting, we consider pertinent negatives as they are intuitively more similar, and therefore comparable, to other counterfactual-based explainability methods. For each generated pertinent negative, we calculate the importance score for each feature by multiplying the absolute change from the value in the original instance to the value in the pertinent negative, modeled as the standard deviation ($SD$) of that feature $\Tilde{X}({h_s})$ across all instances used for the experiment ${X}({h_s})$, as shown in the following formula:
\begin{equation}
\text{CEM}({h_s}) = [X({h_s}) - \Tilde{X}({h_s})]\times SD(h_s)
\end{equation}
The importance score therefore takes into consideration both the necessary perturbation of the feature as well as the significance of the change relative to the feature range. We normalize the scores in the range $[0,1]$, such that the resulting feature importance weights $e_s$ can be directly comparable.

\textbf{Diverse Counterfactual Explanations (DiCE)} \cite{dice} generates example instances to explain the model prediction as well. However, while CEM describes conditions necessary to keep the prediction unchanged, DiCE describes the smallest possible change to the initial instance that results in a different prediction. In other words, DiCE generates nearest neighbor counterfactual examples by optimizing the loss:
\begin{equation}
    \begin{aligned}
    \text{DiCE}({h_s})=\underset{\boldsymbol{c}_{1}, \ldots, \boldsymbol{c}_{k}}{\arg \min } & \frac{1}{k} \sum_{i=1}^{k} \operatorname{y_{loss}}\left(g\left(\boldsymbol{c}_{i}\right), y\right) \\
    &+\frac{\lambda_{1}}{k} \sum_{i=1}^{k} \operatorname{dist}\left(\boldsymbol{c}_{i}, {h_s}\right) \\
    &-\lambda_{2} \operatorname{ diversity}\left(\boldsymbol{c}_{1}, \ldots, \boldsymbol{c}_{k}\right)
    \end{aligned}
\end{equation}
where $\boldsymbol{c}_{i}$ is a counterfactual example, $k$ is the total number of examples to be generated, $g$ is the black box ML model, $y_{loss}$ is a metric that minimizes the distance between the prediction $g^{\prime}$ makes for $\boldsymbol{c}_{i}$ and the desired outcome $y$, $h_s$ is the original input with $|h_s|$ input features, and diversity is the Determinantal Point Process (DPP) diversity metric. $\lambda_{1}$ and $\lambda_{2}$ are hyperparameters that balance the three parts of the loss function. The stopping condition is convergence or 5000 time steps per counterfactual. Microsoft's DiCE library \cite{dice} has a built-in function to compute local feature importance scores from the counterfactual instances, scaled in $[0,1]$. We use them as feature importance weights $e_s$.

\begin{table*}[]
\small
\resizebox{\textwidth}{!}{
\begin{tabular}{lllllrrrr}
\toprule
\textbf{Title} & \textbf{Identifier} & \textbf{Topic$^1$} & \textbf{Level} & \textbf{Language} & \multicolumn{1}{r}{\textbf{\begin{tabular}[c]{@{}c@{}}No. \\ Weeks\end{tabular}}} & \multicolumn{1}{r}{\textbf{\begin{tabular}[c]{@{}c@{}}No. \\ Students$^2$\end{tabular}}} & \multicolumn{1}{r}{\textbf{\begin{tabular}[c]{@{}c@{}}Passing \\ Rate (\%)\end{tabular}}} & \multicolumn{1}{r}{\textbf{\begin{tabular}[c]{@{}c@{}}No. \\ Quizzes\end{tabular}}} \\
\midrule
Digital Signal Processing 1 & \textit{DSP 1} & CS & Bsc & French & 10 & 5629 & 26.8 & 17 \\
% \midrule
Digital Signal Processing 2 & \textit{DSP 2} & CS & MSc & English & 10 & 4012 & 23.1 & 19 \\
% \midrule
Éléments de Géomatique & \textit{Geomatique} & Math & MSc & French & 15 & 452 & 45.1 & 27 \\
% \midrule
Villes Africaines & \textit{Villes Africaines} & SS & BSc & English & 13 & 5643 & 9.9 & 17 \\
% \midrule
Comprendre les Microcontrôleurs & \textit{Micro} & Eng & BSc & French & 13 & 4069 & 5.1 & 18 \\
\bottomrule
\end{tabular}}
\footnotesize{$^1$\textbf{Topic abbrev.} \textit{Eng}: Engineering; \textit{Math}: Mathematics; \textit{CS}: Computer Science; \textit{SS}: Social Science}\\
\footnotesize{$^2$\textbf{No. Students} is calculated after filtering out the early-dropout students, as detailed in Sec. \ref{sec:log-preproc}.}
\caption{Detailed information on the five MOOCs included in our experiments.}
\label{tab:courses}
\end{table*}

\section{Experimental Analysis}
\label{sec:results}
We evaluated the explainability methods on five MOOCs. We first explored how feature importance varies across different explainers for one specific course $c$ (\textbf{RQ1}). We then investigated the similarity of the explainability methods across the five courses using distance metrics (\textbf{RQ2}). Finally, we assessed the validity of the explainers using simulated data from a course $c$ with a known underlying prerequisite skill structure (\textbf{RQ3}). In the following sections, we describe the dataset and optimization protocol used for the experiments before explaining each experiment in detail.

\vspace{-1mm}
\subsection{Dataset}

Our experiments are based on log data collected from five MOOCs of École Polytechnique Fédérale de Lausanne between 2013 to 2015. We chose the five courses to cover a diverse range of topic, level, and language. Table \ref{tab:courses} describes the five courses in detail. We include two subsequent iterations of the same computer science course (\textit{DSP}) with different student populations (French Bachelor students vs. English MSc students). Besides computer science, we also cover courses in the areas of mathematics (\textit{Geomatique}), social sciences (\textit{Villes Africaines}) and engineering (\textit{Micro}). In total, the raw data set contained log data from 75,992 students. After removing the early-dropout students (see Sec. \ref{sec:log-preproc}), 19,805 students remain in the data set. The smallest course contains $452$ students, while the largest course contains 5,643 students. Students' log data consists of fine-grained video (e.g., play, pause, forward, seek) and quiz events (e.g., submit). Interaction data is fully anonymized with regards to student information, respecting participants' privacy rights.

\vspace{-1mm}
\subsection{Experimental Protocol}
\label{sec:opt-proc}
For each course $c \in C$, we trained a \emph{BiLSTM} model $M_c$ on features $H_c$ extracted from $c$. For the optimization, we used batches of size $32$, an Adam optimizer with an initial learning rate of $0.001$, and a binary crossentropy loss. After an initial grid search\footnote{Grid search is discussed further in Appendix A.}, we selected the same architecture for all models: two \emph{BiLSTM} layers consisting of {64, 32} units and one Dense layer consisting of {1} unit with a Sigmoid activation. As this work is not focused on improving model performance, we did not tune hyperparameters further. Formally, we split the data of each course $c$ into a training data set $S_{train,c}$ ($80\%$ of the students) and a test data set $S_{test,c}$  ($20\%$ of the students). For each course, we performed a stratified train-test split over students' pass/fail label. We then trained each model $M_c$ on the training data set $S_{train,c}$ and then predicted student success on the respective test data set $S_{test,c}$. We chose the balanced accuracy (BAC) as our primary evaluation metric because of the high class imbalance of most of the selected courses.

For the first two experiments (\textbf{RQ1} and \textbf{RQ2}) we used the student log data collected for the full duration of the course for training and prediction of our models. In the third experiment, we optimized models for different sequence lengths, i.e. using only the log data up to a specific week $w$ of the course (i.e. from week $1$ to week $w$) to predict performance in the assignment of course week $w$. Additional replication details for model training can be found in Appendix B.

For all experiments, we applied the explainability methods to the predictions of the optimized models $M_c$. All five methods are instance-based; they compute the feature importance based on the model predictions for a specific instance. Training explainers on the scale of thousands of students across five courses is not feasible due to the computation time required to generate the explanation for one instance (e.g., the counterfactual explainability methods take a computation time of 30 minutes per instance $s$). Therefore, we determined a representative sampling strategy to pick $100$ students from each course $c$, resulting in explanations for $500$ students in total\footnote{Sampling strategy is discussed further in Appendix C.}. For the first two experiments (\textbf{RQ1} and \textbf{RQ2}), we used a uniform sampling strategy to select the representative students $s^{r}_{i,c}$ for a course $c$ and ensured balance between classes (pass/fail). We first extracted all failing students and ordered them according to the predicted probability of the model $\hat{p}(l_{S_i}=0)$. We then uniformly sampled $50$ failing students from this ordered interval. We repeated this exact same procedure to sample the $50$ passing students. This sampling procedure ensures that we include instances where the model is confident and wrong, instances for which the model is unsure, and instances where the model is confident and correct. For the last experiment (\textbf{RQ3}), we used performance in the assignment of a given week $w$ as the binary outcome variable. We then followed exactly the same uniform sampling procedure as for \textbf{RQ1} and \textbf{RQ2}, ensuring class balance on assignment performance.

\begin{figure*}[t!]
\centering
\includegraphics[width=\linewidth, trim=4 4 4 4,clip]{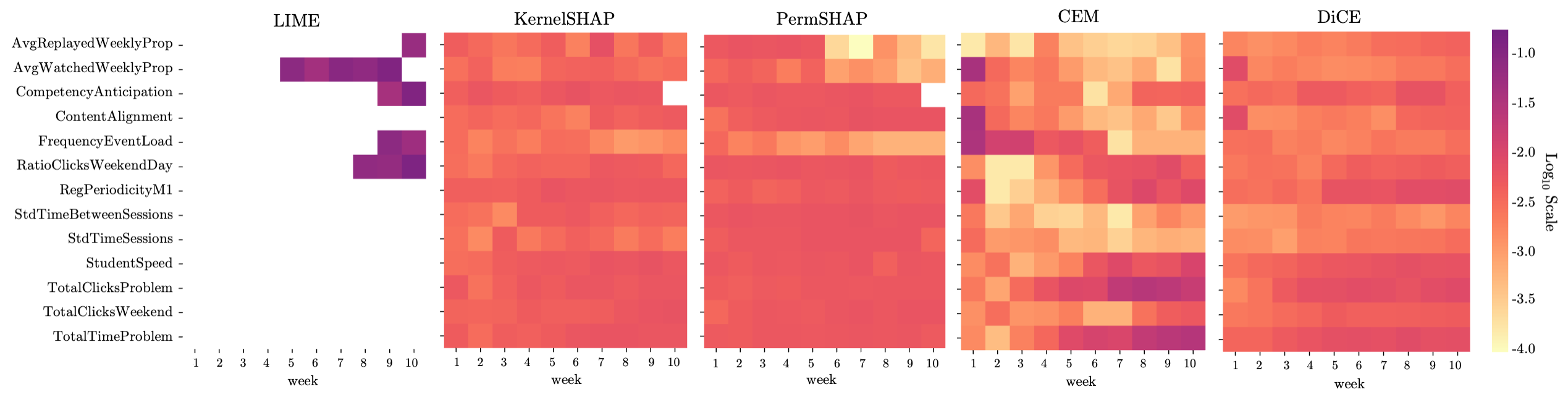}
\caption{Heatmap of normalized feature importance scores (log scale) across explainability methods for \textit{DSP 1}.}
\label{fig:feature_importance_heatmaps}
\end{figure*}

\vspace{-2mm}
\subsection{RQ1: Explanations for one course}
\label{subsec:rq1}
In a first experiment, we compared the explanations of the instance-based methods for one specific course (\textit{DSP 1}). The \emph{BiLSTM} model $M_{DSP1}$ trained on this course achieved a BAC of $93.9\%$. We then ran the explainability methods % (LIME, KernelSHAP, PermSHAP, DiCE, CEM)
on $M_{DSP1}$ and extracted normalized feature importance scores for $100$ representative students of each course.

Figure \ref{fig:feature_importance_heatmaps} illustrates the features identified as most important by each explainability method. The heatmaps were computed by averaging importance scores for each feature and week across $100$ representative students for \textit{DSP 1} (see Sec. \ref{sec:opt-proc}). To ensure interpretability of Figure \ref{fig:feature_importance_heatmaps}, we only included the top five features for each method, resulting in $13$ distinct features. The description of all the features can be found in Table \ref{tab:features}. We used a log scale within the heatmaps, with darker colors indicating higher feature importance.

We observe that the top features cover all the different behavioral aspects included in the feature set: Regularity, Engagement, Control, and Participation. However, some aspects seem to contain more important features. For example, $43\%$ of the Participation features ($3$ out of $7$ features) are in the top five features of at least one method, while this is the case for only $23\%$ ($5$ out of $22$ features) of the Control features. For Regularity and Engagement, $33\%$ and $31\%$ of the features get selected into the top feature set.

We also immediately recognize that the heatmap of LIME looks very different from the heatmaps of all the other methods. LIME assigned high importance scores to a small subset of features and weeks, while all the other explainability methods tend to identify more features and weeks as important, resulting in generally lower importance scores. We also observe that LIME does not consider student behavior in the first weeks of the course important; all importance is placed onto the second half of the course. Moreover, LIME seems to put more emphasis on Control than on the other three aspects: the features related to Control (\textit{AvgReplayedWeeklyProp}, \textit{AvgWatchedWeeklyProp}, \textit{FrequencyEventLoad}) are important from week $5$ through week $10$, while the features related to Participation (\textit{CompetencyAnticipation}) and Engagement (\textit{RatioClicksWeekendDay}) are important only during the last $2$ to $3$ weeks of the course.

Interestingly, while CEM and DiCE are both counterfactual methods, their heatmaps look quite different: the feature importance scores of DiCE tend to be more similar to KernelSHAP and PermSHAP than CEM. We note that CEM shows a higher diversity in feature importance scores than the other three methods (KernelSHAP, PermSHAP, and DiCE), for which the importance values seem to be quite equally distributed across the top features. Furthermore, in contrast to all the other explainability methods, CEM seems to also identify features in the first weeks of the course as important (e.g., \textit{AvgWatchedWeeklyProp}, \textit{ContentAlignment}, and \textit{FrequencyEventLoad} in week $1$). In contrast to all the other methods, CEM identifies features related to being engaged in quizzes as relevant (\textit{TotalClicksProblem} and \textit{TotalClicksWeekend}). Finally, as expected, the heatmaps of KernelSHAP and PermSHAP look very similar, with only small differences in importance scores.

\textit{In summary, while there is some agreement on the top features across explainability methods (the union of the top five features of each method only contains $13$ distinct features), we observe differences across methods when it comes to exact importance scores.}

\vspace{-1mm}
\subsection{RQ2: Comparing methods across courses}
\label{sec:cross_examine}
Our second analysis had the goal to \emph{quantitatively} compare the explanations of the different methods across all five courses. Explainability method evaluation is an emerging field; most existing research focused on assessing the quality of explanations \cite{sokol2020explainability, leake2014evaluating} with only few works suggesting a quantitative `goodness' score for each explainability method (e.g., \cite{nguyen2020quantitative, zhou2021evaluating}). In contrast, we examined the distance between the feature importance scores per explainability method in comparison to each other, instead of individually. We first visualized the similarity of importances across courses using a Principal Component Analysis and then computed Spearman's Rank-Order Correlation as well as Jensen-Shannon Distance to assess similarity regarding the feature importance ranking as well as their exact values.

\begin{figure}[!b]
\centering
\includegraphics[width=\linewidth, trim=4 4 4 4,clip]{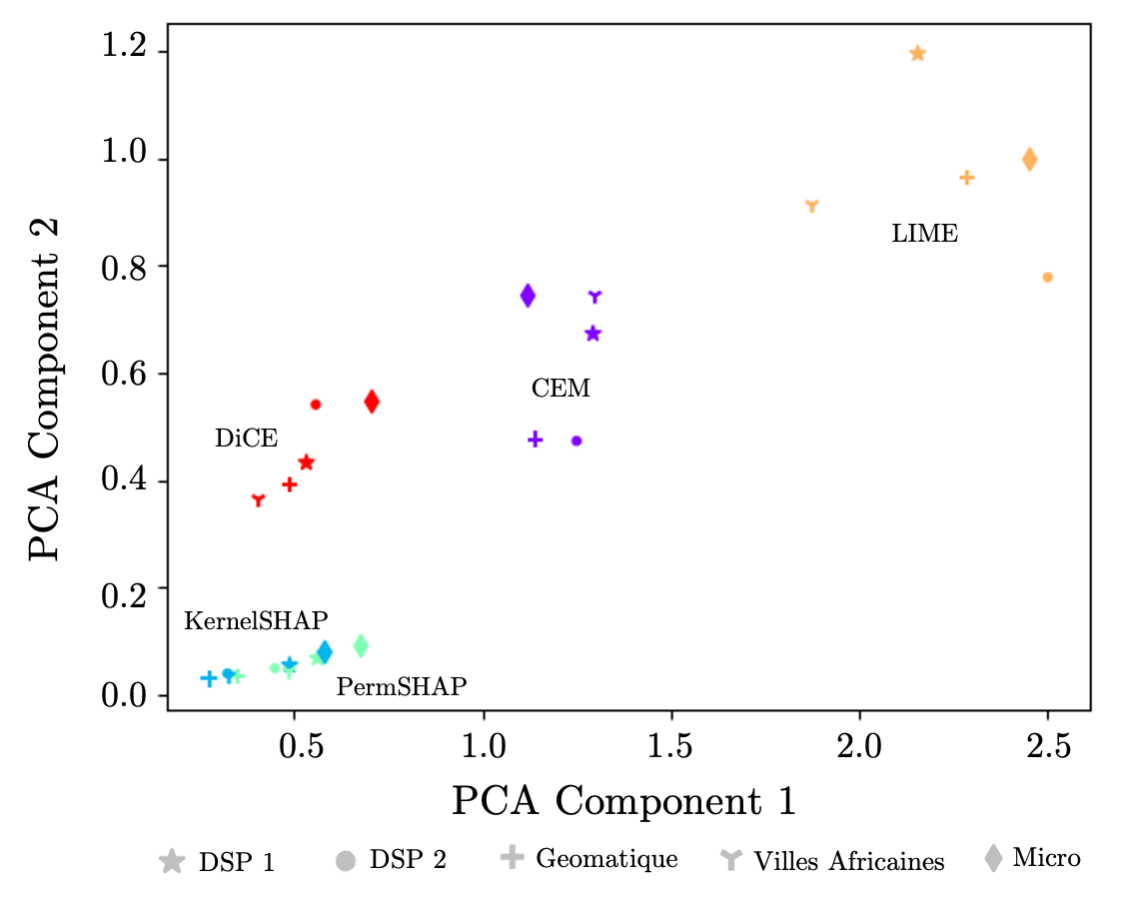}
\caption{PCA of feature importance scores for five explainability methods across five courses.}
\label{fig:pca_all}
\end{figure}

\vspace{1mm} \noindent \textbf{Principal Component Analysis (PCA)} We performed a PCA on the importance scores for each feature and week (length: $w_{c}\times h$) separately for each explainability method and course $c$. Figure \ref{fig:pca_all} shows the results for all explainability methods and courses. Each marker in Figure \ref{fig:pca_all} represents a specific course, while each color denotes an explainability method.

\begin{figure*}[h!]
\centering
\includegraphics[width=\linewidth, trim=4 4 4 4,clip]{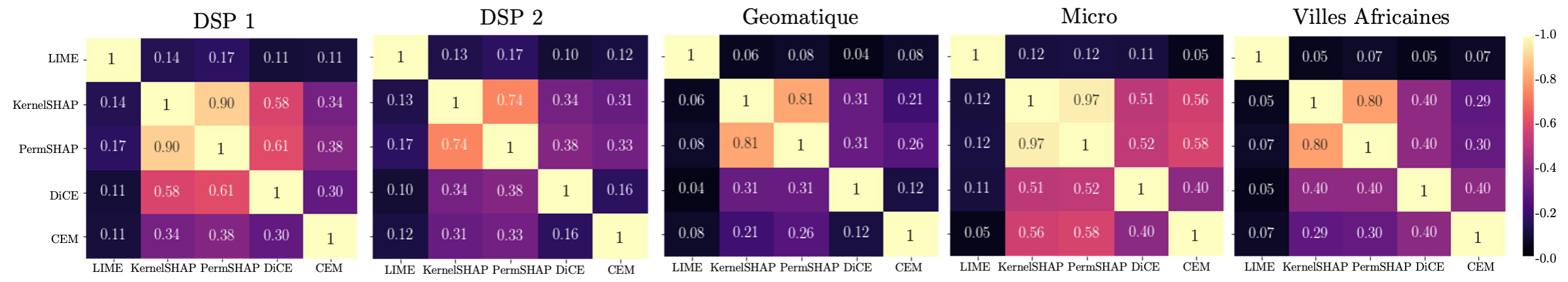}
\caption{Comparison of feature importance scores across courses using Spearman's Rank-Order Correlation.}
\label{fig:spearman}
\end{figure*}

\begin{figure*}[t]
\centering
\includegraphics[width=\linewidth, trim=4 4 4 4,clip]{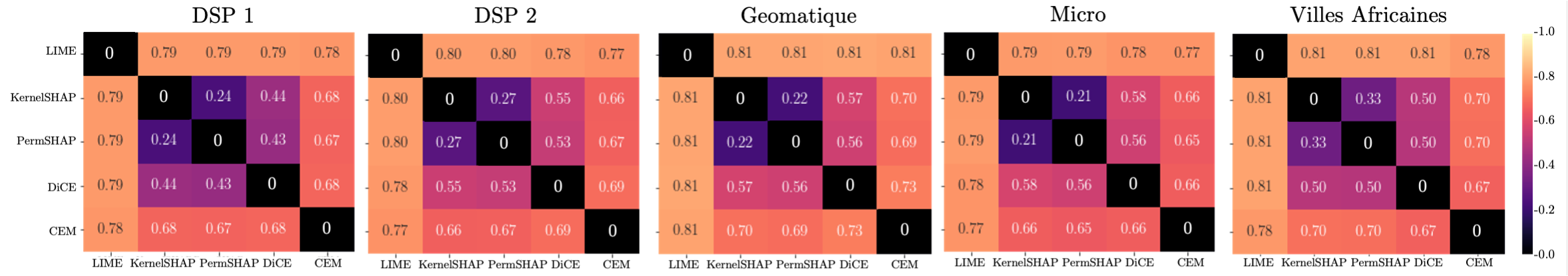}
\caption{Comparison of feature importance scores across courses using Jensen-Shannon Distance.}
\label{fig:jensen_shannon}
\end{figure*}

We observe that the two SHAP methods (KernelSHAP and PermSHAP) cluster together very strongly. This result is expected, as the methodologies of KernelSHAP and PermSHAP are very similar. DiCE feature importances are quite close to the SHAP methods, showing that the three methods have similar notions of feature importance. LIME is quite different from all other methods, with high values on both PCA components. Based on differences in methodology, we would have expected that the difference between the counterfactual methods (DiCE and CEM) and the SHAP methods would be larger than the difference between LIME and the SHAP methods. The most notable takeaway from Figure \ref{fig:pca_all} is that there are clearly identifiable clusters based on explainability method and not on course. It therefore seems that the resulting feature importance scores are mainly influenced by the explainability rather than by the model or data (i.e. the characteristics of the course and students' data).

\vspace{1mm} \noindent \textbf{Spearman's Rank-Order Correlation.} Often referred to as Spearman's $\rho$ \cite{spearman1961proof}, this metric identifies the rank correlation (statistical dependence between the rankings) between two variables and is defined as the Pearson correlation coefficient between the rankings of two variables. We chose this metric for evaluating explainability methods to highlight the importance of feature ranking order in explanations.
To compute Spearman's Rank-Order Correlation $r_{m_1,m_2,c}$ between two explainability methods $m_1$ and $m_2$ on a course $c$, we first converted the vectors $e_{m_1,s}$ and $e_{m_2,s}$ of feature importance scores (length $w_c \times h$) for each student $s$ into rankings $R(e_{m_1,s})$ and $R(e_{m_1,s})$. We then computed $r^{s}_{m_1,m_2,c}$ separately for each relevant student $s$ and then averaged over all relevant students to obtain $r_{m_1,m_2,c}$.

Figure \ref{fig:spearman} illustrates the pairwise similarities between explainability methods using Spearman's Rank-Order Correlation. Higher values imply stronger correlation between methods. We see similarities between KernelSHAP and PermSHAP prevalent once again as a center square for each course, affirming our intuition that two similar methodologies would result in similar rank-order scores. It can be observed that LIME consistently shows rank-order correlation scores with all other explainability methods. Additionally, for \textit{DSP 1} and to some degree also \textit{Villes Africaines}, DiCE is much closer to KernelSHAP and PermSHAP then to CEM. For \textit{DSP 2} and \textit{Geomatique}, DiCE and CEM are both equally correlated to the SHAP methods, but less correlated among themselves. Finally, the model trained on \textit{Micro} has strong correlations across all explainability methods except LIME.

\vspace{1mm} \noindent \textbf{Jensen-Shannon Distance.} We used the Jensen-Shannon distance \cite{jensenshannondivergence} to compute pairwise distances between exact feature importance score distributions obtained with different explainability methods. The Jensen–Shannon distance is the square root of the Jensen-Shannon divergence, originally based on the Kullback–Leibler divergence with smoothed values. It is also known as the Information Radius (IRad) \cite{manning1999foundations}. To compute the Jensen-Shannon distance $jsd_{m_1,m_2,c}$ between two explainability methods $m_1$ and $m_2$ on a course $c$, we first calculated the distance $jsd^{s}_{m_1,m_2,c}$ between the feature importance scores (length $w_c \times h$) $e_{m_1,s}$ and $e_{m_2,s}$ separately for each representative student $s$ and then averaged across all representative students to obtain $jsd_{m_1,m_2,c}$.

Figure \ref{fig:jensen_shannon} shows the pairwise distance between explainability methods for all courses using Jensen-Shannon Distance. Larger numbers represent higher dissimilarity. The Jensen-Shannon Distance heatmaps confirm the observations made using Spearman's Rank-Order Correlation (see Figure \ref{fig:spearman}). Again, LIME consistently has a high distance to all other explainability methods across all courses. As expected, KernelSHAP and PermSHAP have low pairwise distances for all courses. However, when comparing feature importance scores directly instead of using rankings, we observe even less differences between courses. DiCE is closer across all courses to the SHAP methods than CEM. While LIME exhibits the highest distances to all other explainability methods, the explanations of CEM are also far away from all methods.

\textit{In summary, the two SHAP methods and DiCE seem to deliver the most similar explanations, while the feature importance scores obtained with CEM and LIME are different from the other explainability methods. More importantly, all our analyses (PCA, Spearman's Rank-Order Correlation, Jensen Shannon Distance) demonstrate that the choice of explainability method has a much larger influence on the obtained feature importance score than the underlying model and data.}

\vspace{-1mm}
\subsection{RQ3: Validation of explanations}

\begin{figure}[]
\centering
\includegraphics[width=\linewidth, trim=4 4 4 4,clip]{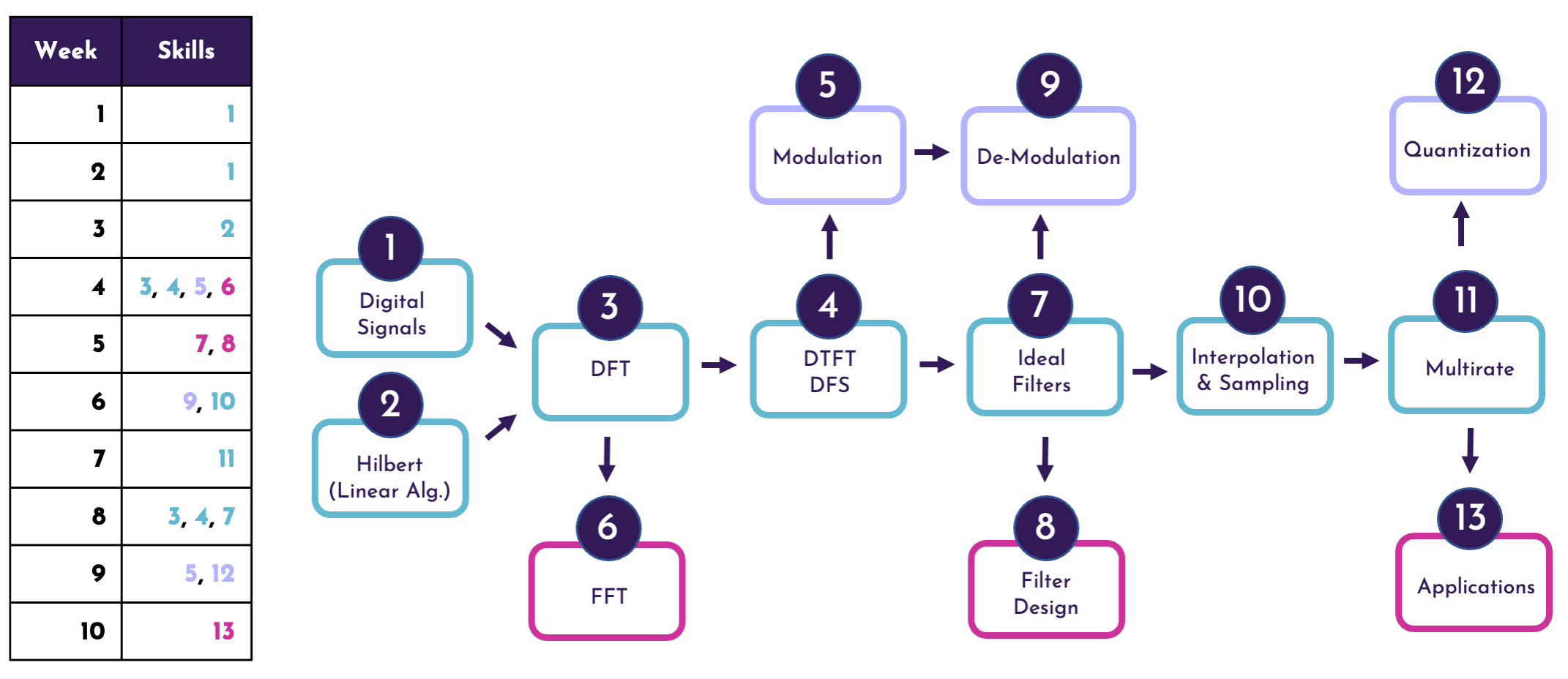}
\caption{Prerequisite skill structure for \textit{DSP 1}.}
\label{fig:skillmap}
\end{figure}

The previous experiments delved into mapping the similarities and differences between explainability methods. While our analyses demonstrated that there are clear disparities across method choice, they do not give an indication regarding the `goodness' of the obtained explanations. Recent work discusses traits of ideal explanations \cite{carvalho2019machine} and targets metrics to measure explanation quality \cite{zhou2021evaluating, jia2021studying}. However, these metrics tend to be over-specialized to one explainability method over others due to the similarities in their methodologies. The community does not yet have a set of standard metrics for evaluating explainability methods. In our last experiment, we hence use information inherent to our model's setting to perform an initial validation of the explanations provided by the different methods.

Specifically, we evaluated the explainability methods on a course with a known underlying skill map and used the prerequisite relationships between weeks of the course as a ground truth for the explanations. Based on the results obtained for the first two research questions (Sections \ref{subsec:rq1} and \ref{sec:cross_examine}), we selected one representative method from each methodological group for the analysis, while keeping the observed explanation diversity: PermSHAP (the most widely used SHAP method), CEM (chosen as a representative of counterfactuals for its further disparity from the SHAP methods), and LIME. In terms of courses, we used \textit{DSP 1} as a basis for the analysis as the instructor of this course provided us with the skill map derived from the curriculum.

Figure \ref{fig:skillmap} illustrates the underlying skills, their relationships, as well as their mapping to the weeks of the course. The arrows denote the prerequisite relationships, while the numbers denote the unique skills in the order they are introduced in the course. The skills colored in pink ($6$, $8$, and $13$) refer to applied skills learned in the course. The middle track refers to core skills learned in the course (colored in blue) and the purple skills at the top ($5$, $9$, and $11$) are theory-based extensions of core material. The skill prerequisite map allows us to analyze the dependencies between the different weeks of the course. For example, in order to understand \textit{Modulation} taught in week $5$, students need to already have learned the skills taught in weeks $3$ and $4$ (\textit{DFT}, \textit{DTFT}, and \textit{DFS}). Intuitively, a model predicting performances in assignments in week $5$ would have highly correlated features based on week $4$. We assume that these dependencies would logically be uncovered by the explainability method.

We, therefore, adjusted our predictive task: using the optimization protocol and experimental design described in Section \ref{sec:opt-proc}, we aimed at predicting the performance (binary label: below average or above average) of a student $s$ in the assignment of week $w$ based on features extracted from student interactions for weeks $1$ to $w$. Given the prerequisite structure for the course, we ran experiments for $w \in \{5, 9\}$. For each predictive model $M_{DSP1,w}$, we then picked the $100$ representative students using uniform sampling and taking into account class balance (see Section \ref{sec:opt-proc} for a detailed description of the sampling procedure) and applied the selected explainability methods to these representative instances.

\begin{figure}[]
\centering
\includegraphics[width=1.05\linewidth, trim=4 4 4 4,clip]{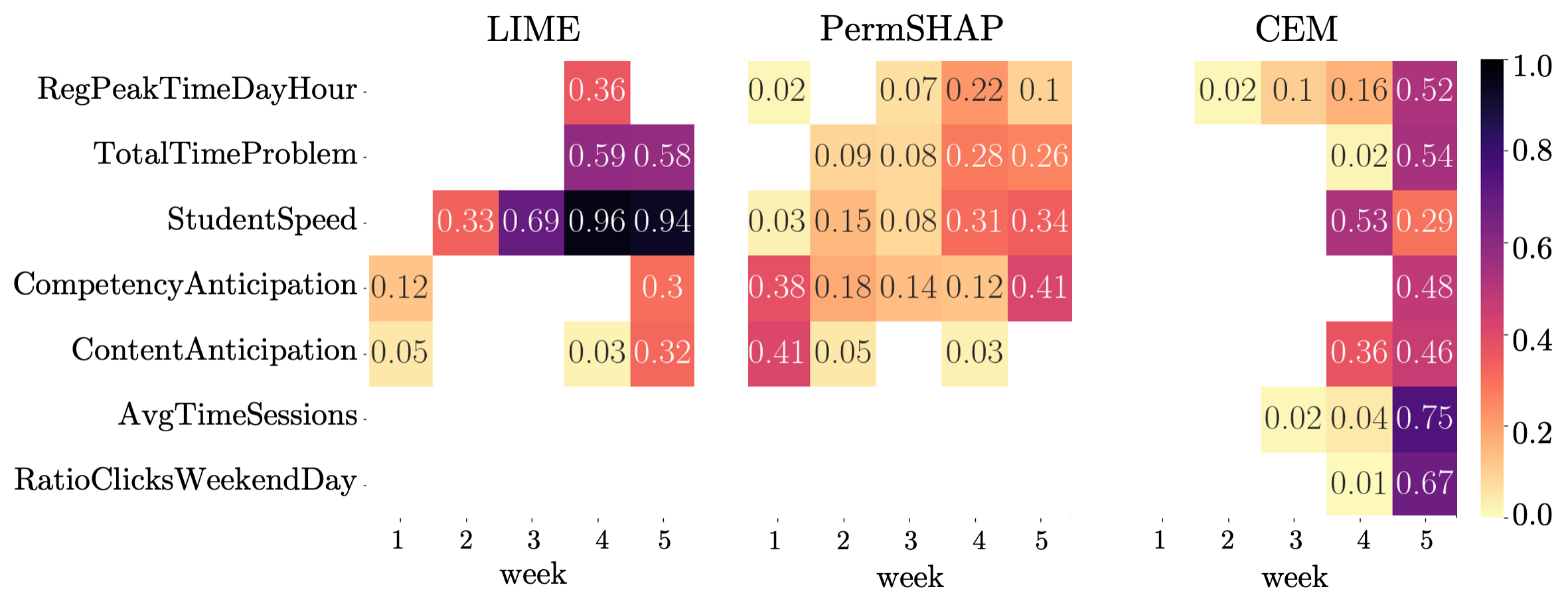}
\caption{Importance scores for LIME, PermSHAP, and CEM for week $5$.}
\label{fig:week5_sim}
\end{figure}

Figure \ref{fig:week5_sim} shows the features for LIME, PermSHAP, and CEM for the prediction model of week $5$. In the heatmap, darker values indicate a higher score. The scores for the heatmap have been computed based on a ranking of features and weeks: for each student $s^{r}_{i}$, we first ranked the features in order of feature importance as determined by the respective explainability method. We then scored each of the top $10$ features according to its rank: $10$ points for the top feature, $9$ points for the second most important feature, and so on. Finally, we averaged the scores for each feature across the $100$ representative students $s^{r}_{1}$ through $s^{r}_{100}$ and normalized them. This rank-based scoring allows us to compare explainability methods without having the relative feature importance scores bias the analysis. We only selected features with a score of at least $0.33$ in any course week, showing only the top two-thirds of features per method.

\begin{figure*}[]
\centering
\includegraphics[width=\linewidth, trim=4 4 4 4,clip]{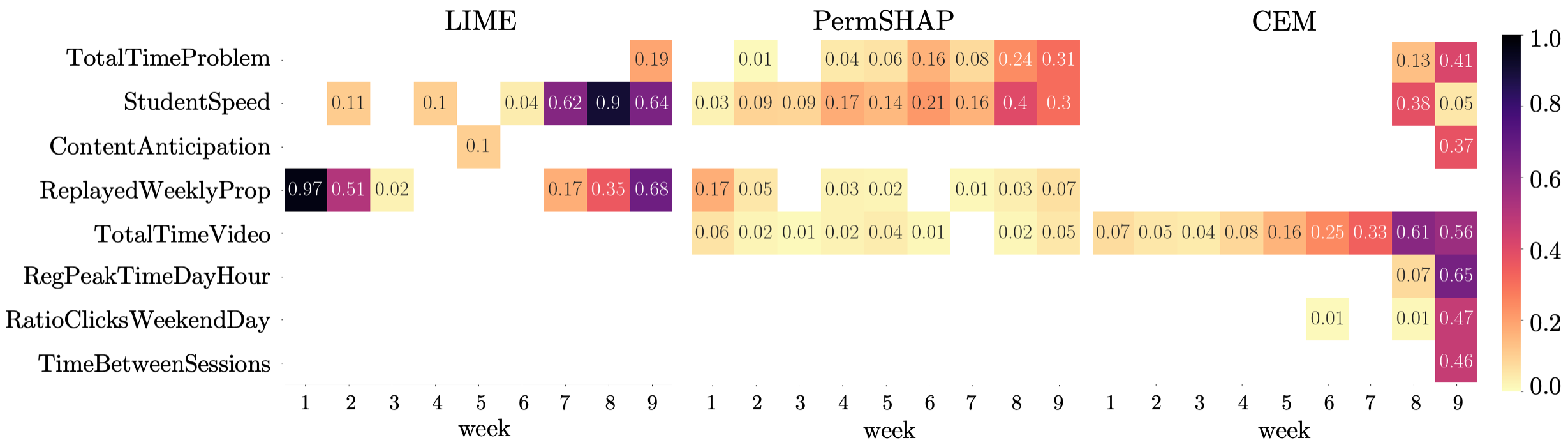}
\caption{Importance scores for LIME, PermSHAP, and CEM for week $9$.}
\label{fig:week9_sim}
\end{figure*}

For LIME, we observe that four features have been identified as important for predicting assignment performance in week $5$. Two of these features directly relate to student behavior on the week $5$ assignment: \textit{TotalTimeProblem} (the total time the student spent solving the assignment) and \textit{StudentSpeed} (the time between consecutive trials of assignments). The high scores of these features for week $5$ are thus expected. LIME also assigns high scores to these two assignments based features for week $4$. This result is encouraging as week $4$ teaches the prerequisite skills of week $5$ and therefore, LIME seems to (at least partially) uncover the prerequisite structure of the course.

In PermSHAP results, we see again that the scores are more uniformly distributed across features and weeks. However, we observe again that the assignment-based features (\textit{TotalTimeProblem} and \textit{StudentSpeed}) have comparably high scores for week $4$ and therefore PermSHAP also seems to (partially) uncover the prerequisite relationships. Curiously, watching videos and solving quizzes scheduled for subsequent weeks (\textit{CompetencyAnticipation} and \textit{ContentAnticipation}) is also considered important, hinting that being proactive when learning increases learning success.

CEM seems to be able to partially uncover the prerequisite relationship between weeks as well. For week $4$, one feature related to assignment behavior (\textit{StudentSpeed}) exhibits a high score. Additionally, watching content of the subsequent week (in this case, week $4$'s \textit{ContentAnticipation} for week $5$ material) is important for assignment performance. Otherwise, CEM mainly explains performance in the assignment of week $5$ with student behavior in week $5$: besides the assignment-related features, the students' actions in the inference week are considered important.

Figure \ref{fig:week9_sim} shows the importance scores for LIME, PermSHAP, and CEM for the prediction model of week $9$. Again, darker colors in the heatmaps indicate higher scores. The scores in the heatmaps were computed using the same ranking-based procedure as for Figure \ref{fig:week5_sim}. In week $9$, Figure \ref{fig:skillmap} indicates that weeks $5$ and $6$ cover the prerequisite skills. From Figure \ref{fig:week9_sim}, we observe that LIME does not seem to be able to capture this prerequisite relationship. The top scores for LIME appear in week $9$ itself for an assignment-based feature (\textit{StudentSpeed}) as well as for video control behavior (\textit{AvgReplayedWeeklyProp}, which computes the relative number of video replays). Furthermore, \textit{StudentSpeed} seems to be generally important also in the weeks just before the predicted week (weeks $7$ and $8$). For PermSHAP, we again obtain a more equal distribution, with only the two assignment-related features (\textit{TotalTimeProblem} and \textit{StudentSpeed}) in weeks $8$ and $9$ showing relatively higher scores. Furthermore, PermSHAP also assigns relatively higher importance to these features for week $6$, which is a prerequisite for week $9$. For CEM, we again observe that mainly student behavior in the actual week, i.e. week $9$, seems to explain assignment performance. Only one feature (\textit{TotalTimeVideo}) shows medium importance for weeks $5$ and $6$.

\textit{In summary, all the evaluated methods were able to (partially) detect the prerequisite relationship between week $4$ and week $5$. For week $9$, detecting the prerequisite structure proved to be difficult; results differed between methods. However, we should take into account that none of the features of the feature set directly measure student performance and therefore, the generated explanations rely on behavioral features only. It appears that recent and actual behavior is a much stronger indicator for performance than past behavior.}

\section{Discussion and Conclusion}

Explainability methods allow us to interpret a deep model in a way that is understandable not only to machine learning experts, but also end-users of educational environments, including instructors tailoring course designs and students the model is predicting on \cite{webb2021machine, conati2018ai}. In this paper, we aimed to understand explainers' behaviour and the ways in which they differ for the task of student success prediction.

Our results demonstrate that all explainability methods can derive interpretable motivations behind student success predictions, confirming the similar yet coherent observations made by \cite{lu2020towards} for the knowledge tracing field. However, while there was some agreement regarding the top features across the five explainability methods, key differences across methods emerged when we considered the exact importance scores (\textbf{RQ1}). We observed substantial similarities between KernelSHAP, PermSHAP, CEM and DiCE with regards to the top ranked feature-weeks. Conversely, LIME only ranked very few features as important, and these less important feature similarities made the other explainability methods appear closer to each other. Overall, looking beyond top ranked features, we noted considerable differences in feature importances across explainability methods. Interestingly, LIME-detected features are more in line with the features marked as important by Random Forests in \cite{marras2021can}, still in a MOOC context. This observation further demonstrates the generalizability of the features' predictive power even among very different experimental settings.

In a subsequent experiment, we compared the different explainability methods across five MOOCs. Our findings indicate that the choice of explainability method has a much larger influence on the obtained feature importance score than the underlying model and data (\textbf{RQ2}). With distance (Jensen-Shannon distance) and ranking-based metrics (Spearman's Rank-Order Correlation), we uncovered that LIME is farthest from the other explainability methods. The sparsity of LIME-detected important features was also observed by \cite{scheers2021interactive}, where the conciseness of LIME explanations supported integration in visual dashboards for student advising. We also detected a close relationship between KernelSHAP and PermSHAP, which strongly validates our evaluation strategy. Using PCA, we identified clear clusters of explanations by explainability method and not by the course the model was trained upon, suggesting that an explainability method might be prone to mark specific features as important regardless of the model (and the course).

Our analyses also confirmed that all the evaluated methods were able to (partially) detect the prerequisite relationship between weeks, while relying on behavioral features only (\textbf{RQ3}). Our experimental design was inspired by \cite{mu2020towards}'s work on predicting effectiveness of interventions for wheel-spinning students by simulating prerequisite relationships. While we have no way to examine the true underlying feature importances of our week $n$ assignment performance prediction model, we intuit that a student's prerequisite week performance should be important to predicting their performance in week $n$. We observed that the three families of methods (LIME, SHAP, and counterfactual) were able to partially capture the prerequisite relationship in week $5$, but struggled to capture the prerequisite relationships in week $9$. While there were few similarities in the top ranked features, each method found different groups of feature-weeks as most important for the same models. Our results indicate that recent and current behavior is more important than past behavior, implying that proximity of behavioural features correlates strongly with their perceived importance. A limitation is that the prerequisite relationships we deem important might not actually be used as the true features of the model since our feature set included only features that examined student behavior and not direct performance.

Our results indicate that there are noteworthy differences in generated explanations for student performance prediction models. However, our analyses also show that these explanations often recognize prerequisite-based relationships between features. That being said, our study still has several limitations that warrant future research, including our focus on a singular downstream task (student success prediction), specific modality of dataset (MOOCs), choice of model architecture (\textit{BiLSTM}), and lack of assessment of the obtained explanations' impact in the real world. First, extending our experiments beyond success prediction to a multi-task analysis (e.g., dropout prediction) across multiple modalities (e.g., flipped classrooms, intelligent tutoring systems) would allow us to build stronger intuitions about explainability method differences. Second, extending our black-box BiLSTM model architecture to multiple traditional and deep machine learning architectures could examine whether certain explainability methods have stronger explanation affinity to different predictors. Choosing transparent shallow architectures instead of black-boxes could also allow us to validate our results against ground truth feature importances. Third, further research should be conducted to check which explanations (and explainability methods) lead to interventions that better improve learning outcomes. It follows that an assessment of the obtained explanations should be carried out involving educators. Finally, the disagreement of our selected explainability methods motivate an extension in an ensemble expert-weighting scheme, which might have merit in closely estimating the true feature importances.

Explainability in educational deep learning models can lead to better-informed personalized interventions \cite{xing2019dropout, karimi2021algorithmic}, curriculum personalization, and informed course design. If we were considering global interventions (as it might be too resource intensive to perform interventions on each student individually), we could take the mean feature importance vector over all students and try interventions in the order of the scores of this mean vector. If we were only able to intervene on $k$ features due to resource constraints, Spearman's rank-order metric could also be modified to include the size of the intersection between the features with the top $k$ scores. However, it is important to note that when the model explanations are biased by explainability method and do not accurately reflect the inner workings of the model, the impact of incorrect predictions are further exacerbated by teachers and students' misplaced confidence in the model's justification. \textbf{We implore data scientists to not take the choice of explainability method lightly as it does have a significant impact on model interpretation, and instead urge the community to (1) carefully select an appropriate explainability method based on a downstream task and (2) keep potential biases of the explainer in mind when analyzing interpretability results.} Overall, our work contributes to ongoing research in explainable analytics and to the generalization of theories and patterns in success prediction.

\section{Acknowledgments}
We thank Professor Paolo Pardoni, Valentin Hartmann, and Natasa Tagasovska for their expertise and support.

% The following two commands are all you need in the
% initial runs of your .tex file to
% produce the bibliography for the citations in your paper.
\balance
\bibliography{base}  % sigproc.bib is the name of the Bibliography in this case
% You must have a proper ".bib" file
%  and remember to run:
% latex bibtex latex latex
% to resolve all references
%
%APPENDICES are optional
\appendix
\section{Model Architecture}
We experimented with traditional machine learning models (e.g., Support Vector Machines, Logistic Regression, Random Forest) and deep-learning models (e.g., Fully-Connected Networks, RNNs, LSTMs, CNNs, and BiLSTMs), and found that BiLSTM models perform best against the other baselines for our use case. To determine the optimal model architecture, we evaluate validation set performance on the course \textit{DSP 1} as it is used in all three RQs. BiLSTMs have a $26.8\%$ average increase in balanced accuracy over traditional machine learning methods. For the BiLSTM architecture grid search, we examined the following layer settings \{32, 64, 128, 256, 32-32, 64-32, 128-32, 128-64, 128-128, 256-128, 64-64-32, 128-64-32, 64-64-64-32-32\} before determining 64-32 performed best in balanced accuracy for \textit{DSP 1}. We used the Tensorflow library to train our models \cite{abadi2016tensorflow}.

\section{Model Training}
Model training took approximately 35 minutes per model on an Intel Xeon E5-2680 CPU with 12 cores, 24 threads, and 256 GB RAM. Each model was trained for 15 epochs, and the best performing model checkpoint was saved. The five models' performance metrics are showcased in Table \ref{tab:modelperf}.

\section{Sampling Strategy}
We experimented with several strategies to extract a appropriate representative sample including the greedy algorithm SP-LIME \cite{lime}, random sampling, two sets of extreme students (those which the model predicts very well on and very badly on), and uniform sampling. We determined that our uniform sampling approach was the most fair with respect to the variable class imbalance between courses.

\begin{table}[!b]
\small
\resizebox{0.5\textwidth}{!}{
\begin{tabular}{lrrrr}
\toprule
\textbf{Identifier} & \textbf{Accuracy} & \textbf{Balanced Accuracy} & \textbf{F-1 Score} \\
\midrule
\textit{DSP 1} & 99.3 & 97.4 & 99.6 \\
\textit{DSP 2} & 99.1 & 93.5 & 99.5 \\
\textit{Geomatique} & 97.7 & 96.2 & 98.7 \\
\textit{Villes Africaines} & 98.4 & 95.5 & 99.1 \\
\textit{Micro} & 89.5 & 90.9 & 90 \\
\bottomrule
\end{tabular}}
\caption{Performance of the BiLSTMs trained on the five MOOCs included in Section 3.}
\label{tab:modelperf}
\end{table}

\balancecolumns
% That's all folks!
\end{document}